\documentclass{article}

\usepackage[final]{corl_2017} 

\usepackage{color,soul}
\usepackage{amsmath}
\usepackage{tikz}
\usetikzlibrary{arrows,shapes}
\usepackage{subcaption}
\usepackage{wrapfig}
\usepackage{amsfonts}
\usepackage{verbatim}
\usepackage{authblk}
\usepackage{hyperref}

\def\drawconv[#1,#2,#3,#4] {
    \node[rectangle,fill=red!60,draw,font=\sffamily\tiny,align=center,anchor=west,minimum height=1.0cm,inner sep=0.06cm] (#4) at ({\xoffset+\xstep*#1},\yoffset) {\rotatebox{270}{\parbox[c]{1.45cm}{\centering {\bf #2:}{\it #3}}}}; %
}

\def\drawpool[#1,#2,#3,#4] {
    \node[rectangle,fill=green!60,draw,font=\sffamily\tiny,align=center,anchor=west,minimum height=1.0cm,inner sep=0.06cm] (#4) at ({\xoffset+\xstep*#1},\yoffset) {\rotatebox{270}{\parbox[c]{1.45cm}{\centering {\bf #2:}{\it #3}}}}; %
}

\def\drawfc[#1,#2,#3,#4] {
    \node[rectangle,fill=blue!60,draw,font=\sffamily\tiny,align=center,anchor=west,minimum height=1.5cm,inner sep=0.06cm] (#4) at ({\xoffset+\xstep*#1},\yoffset) {\rotatebox{270}{\parbox[c]{1.45cm}{\centering {\bf #2:}\\{\it #3}}}}; %
}

\makeatletter
\renewcommand*{\thanks}[1]{%
  \footnotemark
  \protected@xdef\@thanks{\@thanks
    \protect\footnotetext[\arabic{footnote}]{#1}}%
}
\makeatother

\title{Learning Robotic Manipulation of Granular Media}

\author[A,B]{Connor Schenck\thanks{Corresponding author: schenckc@cs.washington.edu}\hspace{0.1cm}}
\author[B]{Jonathan Tompson}
\author[A]{Dieter Fox}
\author[B,C]{Sergey Levine}
\affil[A]{University of Washington}
\affil[B]{Google, Inc.}
\affil[C]{University of California Berkeley}

\begin{document}
\maketitle


\begin{abstract}
In this paper, we examine the problem of robotic manipulation of granular media.
We evaluate multiple predictive models used to infer the dynamics of scooping and dumping actions.
These models are evaluated on a task that involves manipulating the media in order to deform it into a desired  shape.
Our best performing model is based on a highly-tailored convolutional network architecture with domain-specific optimizations, which we show accurately models the physical interaction of the robotic scoop with the underlying media.
We empirically demonstrate that explicitly predicting physical mechanics results in a policy that out-performs both a hand-crafted dynamics baseline, and a ``value-network'', which must otherwise implicitly predict the same mechanics in order to produce accurate value estimates.
\end{abstract}



\setlength{\textfloatsep}{0.35\textfloatsep}

\section{Introduction}
	

\setlength\intextsep{0pt}
\begin{wrapfigure}[24]{R}{0.4\textwidth}
    \centering
    \setlength{\fboxsep}{0pt}
    \setlength{\fboxrule}{1pt}
    \setlength{\unitlength}{1.0cm}
    \fbox{\includegraphics[width=0.38\textwidth]{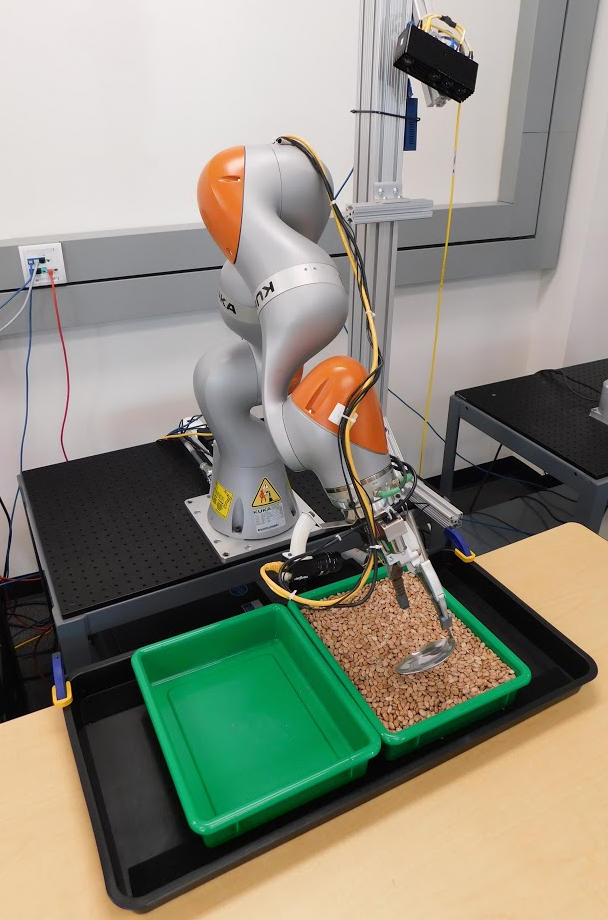}}
    \caption{One of the robots in front of it's table. In the gripper is the rigidly attached spoon used for scooping. On the table is the tray (green) containing the pinto beans used as the granular media. The black tray is to catch spilled beans.}
    \label{fig:robot}
\end{wrapfigure}

Playing in the sand and constructing sand castles is a common pastime for children.
It is easy for a child to view a pile of sand, visualize what they want to make, and construct a reasonable approximation to this desired shape.
These kinds of skills transfer readily to a wide variety of tasks; from manipulating beans while cooking, to using construction machinery to excavate soil.
In fact, it would seem that reasoning about manipulation of granular media and fluids is no more difficult for humans than manipulation of rigid objects.
Yet in robotics there has been relatively little work examining how robots can learn such manipulation skills.
If we want robots to operate intelligently in our environments, then it is necessary for them to be able to reason about substances like sand and beans that do not lend themselves readily to straightforward kinematic modeling.

In this paper, we examine methods for modeling granular media using either a convolutional network (ConvNet) architecture (in both dynamics or value-net variants), or using a fixed-function baseline method.
The goal of this paper is to examine how these models can enable interaction with substances that cannot be modeled easily or accurately with simple kinematics, and whether or not an explicit dynamics prediction is necessary to solve our manipulation task.
To that end, we evaluate the models on a task where the robot must use a scoop to manipulate pinto beans in a tray to form a desired shape (Figure~\ref{fig:robot} shows our experimental setup).
This task domain combines multiple interesting challenges.
Firstly, the beans are difficult to model as a set of discrete rigid objects and thus have non-trivial dynamics.
Secondly, this domain allows for a wide variety of distinct manipulation tasks; such as pushing, collecting, scooping, or dumping to arrange the beans into a desired shape.
Finally, this domain allows us to directly compare multiple models for manipulating granular media, from hard-coded heuristics to learned convolutional networks. 

To solve this task, we develop an algorithm that uses learned predictive models to choose an action sequence which will form a target shape from the granular media.
We study four such predictive models.
\emph{1.} A standard, unstructured ConvNet to predict the shape of the media conditioned on an initial configuration and a candidate manipulation action. \emph{2.} A similar ConvNet model with additional task-specific optimizations built into the architecture to improve prediction accuracy.
\emph{3.} A ConvNet to simply learn whether an action is ``good'' or ``bad'' with respect to a given goal shape.
\emph{4.} A baseline method which uses geometric heuristics to compute a rough estimate of the resulting arrangement of the media, allowing us to measure the value of learning the model rather than specifying it by hand.


We evaluate all of our predictive models on ``easy'' and ``hard'' target shapes.
In both cases, the ConvNets trained to explicitly model the dynamics outperforms other methods.
However, our results show that using a black-box ConvNet does not suffice; rather, policy performance is improved significantly when the network is structured in a way as to preserve essential physical properties of the dynamics.
By using a network trained in a structured manner, the robot is able to accurately predict the state transitions and is better able to choose actions which maximize its potential to reach the goal shape.

\section{Related Work}

In recent years, there has been some work in robotics in areas related to interaction and manipulation of granular media.
For example, there has been a significant amount of work on legged locomotion over granular media \cite{li2009,li2010,li2013}.
There has also been work on automated operation of construction equipment for scooping \cite{hemami1994,takei2013,kanai2006,sarata2004}.
Additionally, much of the work related to robotic pouring has utilized granular media rather than liquids \cite{rozo2013,cakmak2012,yamaguchi2015,yamaguchi2016b,yamaguchi2016c}.
Recent work by Xu and Cakmak \cite{xu2014} explored the ability of robots to clean, among other things, granular media from surfaces using a low-cost tool.
In contrast to this prior work, here we directly tackle the problem of manipulating granular media in a robust, flexible manner.

To manipulate granular media, our robot needs to develop a control policy.
In this paper we focus on policies that utilize ConvNets either for direct policy learning or indirectly (e.g., by estimating dynamics).
While there hasn't been much prior work developing such policies for manipulating granular media, there has been work on developing control policies for tasks such as maze navigation \cite{mirowski2016}, car racing \cite{wierstra2017}, and playing common Atari video games \cite{mnih2013}.
However, all of these were performed in simulators, where it is easy to generate large amounts of data to train the underlying ConvNets.
It is much more difficult to train deep learning algorithms like these in a real robot environment.
Nonetheless, recent work has applied techniques similar to this using methods such as transfer learning from simulation to real-world \cite{sadeghi2016,tzeng2015} or structured learning methods like guided policy search \cite{levine2013,levine2016}.
These types of methods take advantage of other techniques (e.g., simulation or optimal control) to bootstrap their learning models and thus reduce the amount of data needed.
However, an alternative approach is to simply collect more data.
This has been carried out in the literature by utilizing an array of robots, rather than a single robot, and allowing them to collect data continuously over a long period of time \cite{levine2016b,gu2016,finn2016}.
We use this approach in this paper to allow us to collect a large dataset for training our learning models.

One of the main types of models that the robot learns in this paper is a predictive model using ConvNets.
Recent work in robotics has shown how ConvNets can learn pixel-wise predictions for a video prediction task \cite{finn2016},
as well as the dynamics of objects when subjected to robotic poke actions \cite{agrawal2016}.
However, work by Byraven and Fox \cite{byravan2016se3} showed that for dense, unstructured state spaces (in their case raw point clouds from a depth camera), it is often necessary to enforce structure in the network to achieve maximal performance.
In this paper, we compare a standard unstructured network to a structured network for learning dense predictions involving granular media.

Similar to this work, ConvNets have recently been utilized for learning intuition of physical interactions by mimicking physics simulators~\cite{lerer2016learning,chang2016compositional}, for detecting and pouring liquids~\cite{schenckc2016b,schenckc2016c}, and for explicit modeling of rigid body dynamics~\cite{byravan2016se3,wu2016physics} or fluid dynamics~\cite{tompson2016accelerating}. To our knowledge, this work is the first to use ConvNets to predict the dynamics of granular media.

\section{Task Overview}

In this paper the robot is tasked with using a scoop to manipulate granular media in a tray to form a desired shape.
Given an initial state of the environment $h_0$, and a goal state $h_g$, the robot must select a series of actions $a_0, ..., a_T$ that minimizes the L1-norm\footnote{We use the L1-norm rather than Earth mover's distance for the sake of simplicity.} between the final and goal states, i.e.,
\[ l(h_0, a_0, ..., a_T, h_g) = \lVert \mathcal{F}(h_0, a_0, ..., a_T) - h_g \rVert_1 \]
where $\mathcal{F}$ applies the actions $a_0, ..., a_T$ sequentially to the initial state $h_0$ to produce the final state $\mathcal{F}(h_0, a_0, ..., a_T)=h_{T+1}$.
The state $h_t$ is represented as a height-map over the media in the tray and is a 2D grid, where the value in each cell is the height of the surface of the granular media at that location in the tray.
For this paper, each cell is approximately 1 cm $\times$ 1 cm.
An example of a height-map is shown in Figure~\ref{fig:height_map}.

\newlength{\objectsize}
\setlength{\objectsize}{3.3cm}
\begin{figure}
    \centering
    \setlength{\fboxsep}{0pt}
    \setlength{\fboxrule}{1pt}
    \setlength{\unitlength}{1.0cm}
    \begin{subfigure}[b]{\objectsize}
        \fbox{\includegraphics[width=\objectsize]{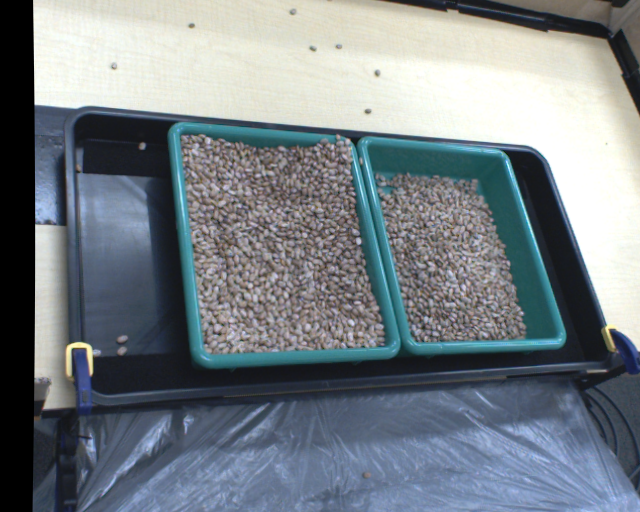}}
        \caption{RGB}
        \label{fig:rgb}
    \end{subfigure}\hspace{0.2cm}%
    \begin{subfigure}[b]{\objectsize}
        \fbox{\includegraphics[width=\objectsize]{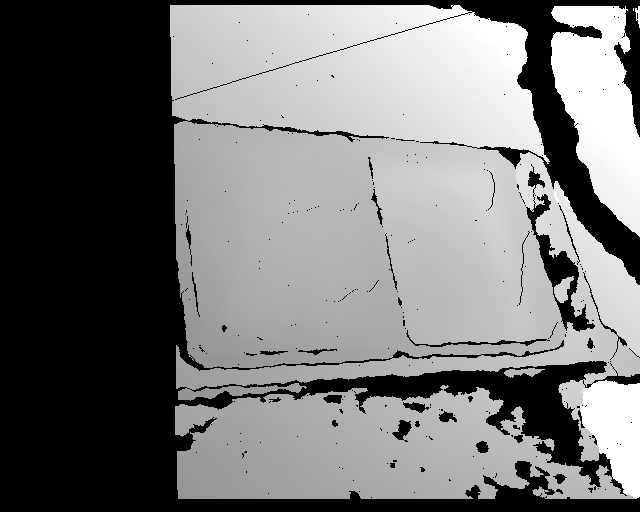}}
        \caption{Depth}
        \label{fig:depth}
    \end{subfigure}\hspace{0.2cm}%
    \begin{subfigure}[b]{\objectsize}
        \fbox{\includegraphics[width=\objectsize]{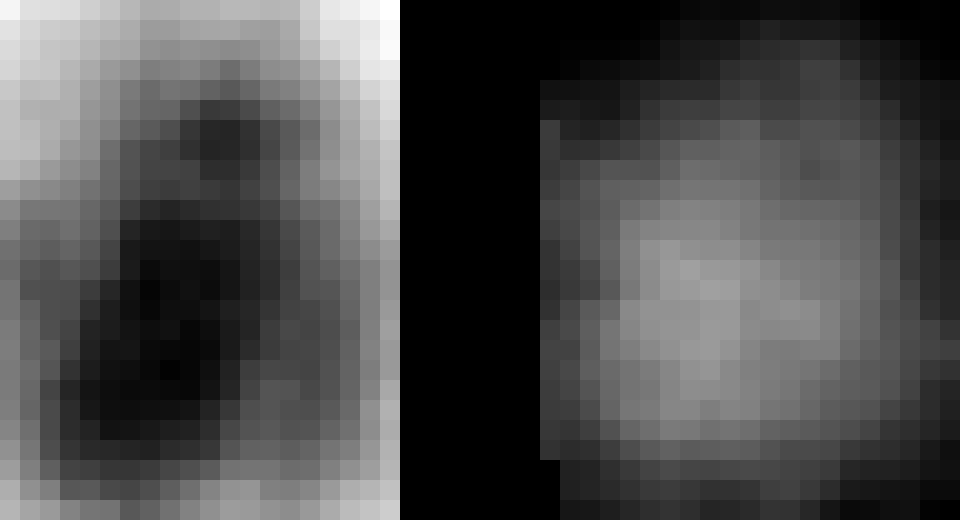}}
        \caption{Height-Map}
        \label{fig:height_map}
    \end{subfigure}\hspace{0.2cm}%
    \begin{subfigure}[b]{\objectsize}
        \fbox{\includegraphics[width=\objectsize]{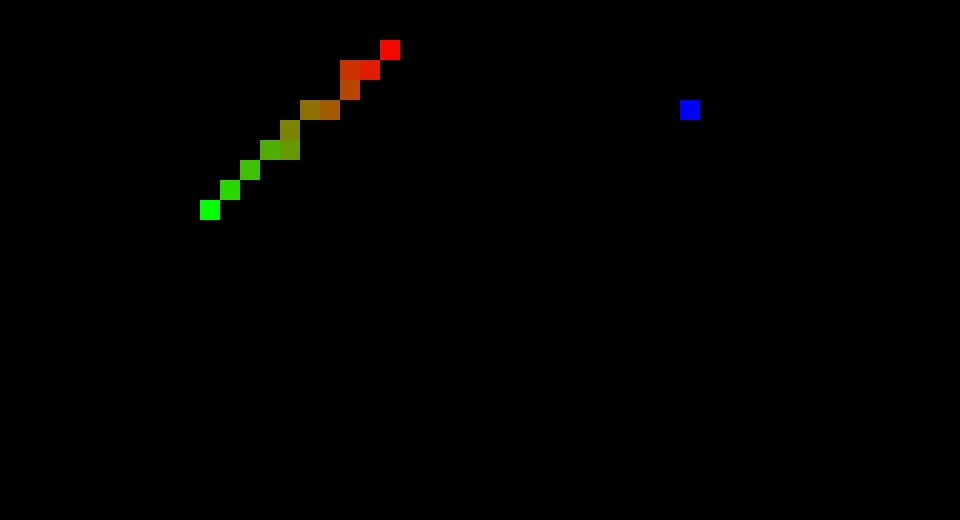}}
        \caption{Action Map}
        \label{fig:action_map}
    \end{subfigure}
    \caption{An example data frame. From left to right: the RGB image, the corresponding depth image, the height-map computed from the depth image, and finally an example action map.}
    \label{fig:data_ex}
\end{figure}

\subsection{The {\it scoop \& dump} Action}
\label{sec:action}

The action the robot performs in this paper is a {\it scoop \& dump} action, which takes a set of parameters $\theta_t$ and then performs a scoop on the bottom of the tray followed by a dump above it.
The parameters $\theta_t$ comprise 6 elements (forming a 9D vector): the start location (2D), the start angle (1D), the end location (2D), the end angle (1D), the roll angle (1D), and the dump location (2D).
The action is split into two parts, {\it scoop} and {\it dump}.
At the start of the {\it scoop}, the robot moves the tip of the scoop to the XY coordinates specified by the start location in $\theta_t$ in the bottom of the tray.
The robot then moves the tip of the scoop in a straight line along the bottom towards the XY coordinates specified by the end location.
During the {\it scoop}, the robot sets the scoop angle, that is the angle between the plane of the tray and the plane of the scoop (with 0 being parallel), by linearly interpolating between the start and end angles specified in $\theta_t$.
Additionally, the robot sets the leading edge of the scoop (the part of the scoop pointing towards the end location) according to the roll angle, with the front leading when the roll angle is 0 degrees and the side of the scoop leading when the roll angle is 90 degrees.
After the scoop reaches the end location, the robot pulls it directly up out of the tray.
The robot then performs the {\it dump}, which consists of moving the scoop tip over the dump location and rotating the scoop to a fixed angle to allow any granular media on it to fall to the tray.

Note that the {\it scoop \& dump} parameterization results in a diverse set of possible behaviors, and can represent complex manipulations in addition to just scooping and dumping; for instance a {\it ``scoop"} with the spoon edge perpendicular to the line of motion will result in a {\it ``pushing"} action, where no beens are collected in the spoon, but are rather deposited at the end of the scoop motion.

We reparameterize the set of action parameters, $\theta_t$, as an ``action map''; an image the same size as the height-map and which encodes the spatial location of our scoop and dump actions.
Starting from an empty image with the same dimensions as the height-map, we draw a straight line from the start location to the end location, and we linearly vary the color of the line from red to green.
Additionally we draw a blue dot at the dump location.
Finally, we tile the 3 angles across 3 extra channels, which we concatenate channel-wise to the action map.
An example of this action map (without the angles) is shown in Figure~\ref{fig:action_map}.
Representing the action parameters in this way brings the actions into pixel-alignment with the ConvNets input state space (the height-map), which makes it more conducive for learning a stable and accurate mapping for our fully-convolutional network architecture (described in the following section).

\section{Predictive Models}

In this paper the robot's control policy that selects the series of actions $a_0, ..., a_T$ relies on an underlying predictive model, i.e., a model that is able to predict some future state based on the current state and a proposed action.
We evaluate four different types of predictive models in this paper.
The first two utilize ConvNets trained to predict the next state given the current state and a proposed action (both describe in the next section).
The third utilizes a ConvNet to predict how much closer (or further) an action will take the robot from a given goal.
The final model we add as a baseline comparison; it uses hard-coded heuristics to predict the next state from the current state and proposed action.
The following sections describe each of these models.
In section~\ref{sec:policy} we describe how the robot utilizes these models to select actions.

\subsection{Predicting the Next State Using ConvNets}
\label{sec:dynamics}

\begin{figure}
\begin{center}
    \begin{tikzpicture}[->,>=stealth,auto,node distance=1.6cm,thick,
      input node/.style={rectangle,draw,anchor=west,align=center,inner sep=0,outer sep=0},
      elt node/.style={ellipse,fill=gray!20,draw,font=\sffamily\LARGE\bfseries,align=center,inner sep=0.04cm}]


	  \pgfmathsetmacro{\xstep}{0.5}
      \pgfmathsetmacro{\xoffset}{1.5}
      \pgfmathsetmacro{\yoffset}{3.975}
      \node[input node] (Ini) at (\xoffset-1.5,\yoffset+0.025) {\includegraphics[width=1.25cm]{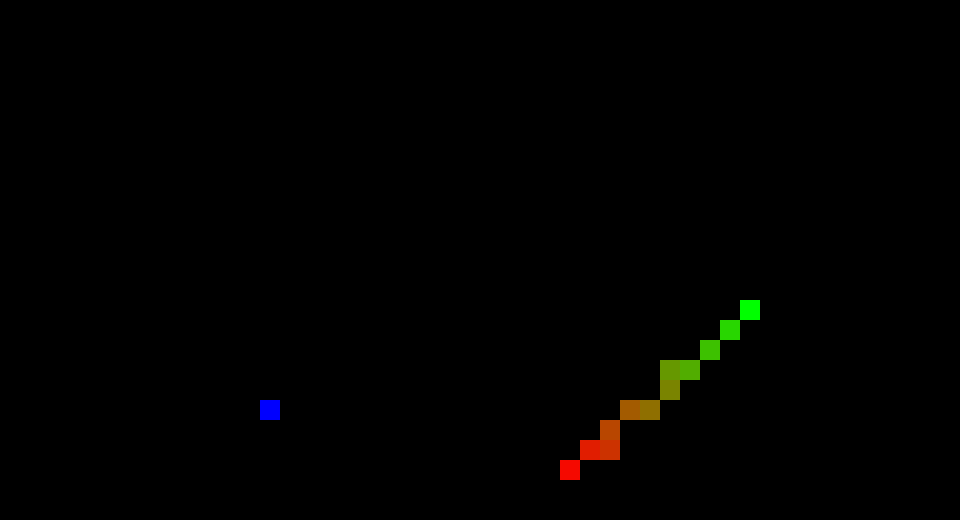}}; 
      \drawconv[0,Conv,32@$3{\times}3$,Conv0i]
      \draw (Ini) -- (Conv0i);
      \drawconv[1,Conv,32@$3{\times}3$,Conv1i]
      \draw (Conv0i) -- (Conv1i);
      \drawconv[2,Conv,32@$3{\times}3$,Conv2i]
      \draw (Conv1i) -- (Conv2i);
      
      \pgfmathsetmacro{\xoffset}{\xoffset-\xstep/3}
      \pgfmathsetmacro{\yoffset}{\yoffset-0.7}
      \node[input node] (Ing) at (\xoffset-1.5,\yoffset+0.025) {\includegraphics[width=1.25cm]{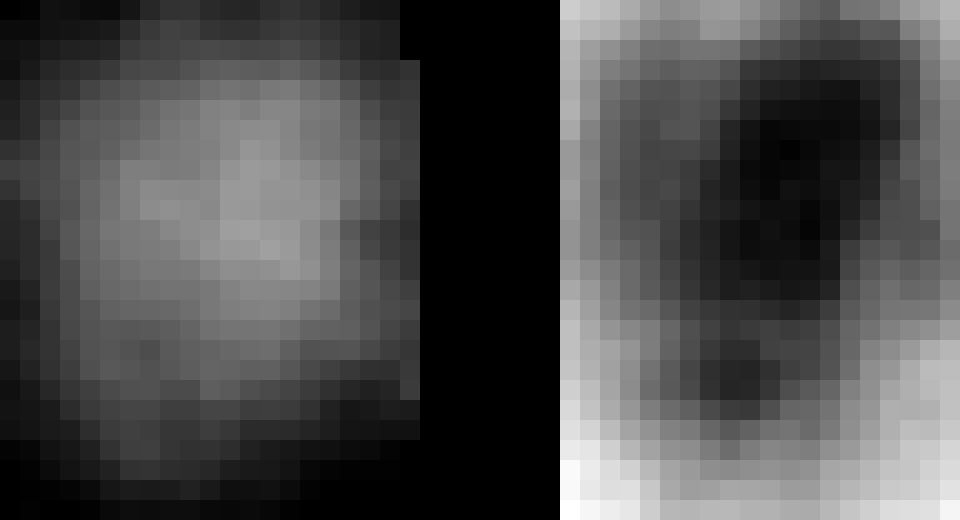}}; 
      \drawconv[0,Conv,32@$3{\times}3$,Conv0g]
      \draw (Ing) -- (Conv0g);
      \drawconv[1,Conv,32@$3{\times}3$,Conv1g]
      \draw (Conv0g) -- (Conv1g);
      \drawconv[2,Conv,32@$3{\times}3$,Conv2g]
      \draw (Conv1g) -- (Conv2g);
      
      \pgfmathsetmacro{\xoffset}{\xoffset+\xstep*4}
      \pgfmathsetmacro{\yoffset}{\yoffset+0.35}
      \drawconv[0,Conv,32@$3{\times}3$,Conv0c]
      \draw (Conv2i) -- (Conv0c);
      \draw (Conv2g) -- (Conv0c);
      \drawconv[1,Conv,32@$3{\times}3$,Conv1c]
      \draw (Conv0c) -- (Conv1c);
      \drawconv[2,Conv,32@$3{\times}3$,Conv2c]
      \draw (Conv1c) -- (Conv2c);
      \drawconv[3,Conv,32@$3{\times}3$,Conv3c]
      \draw (Conv2c) -- (Conv3c);
      \drawconv[4,Conv,32@$3{\times}3$,Conv4c]
      \draw (Conv3c) -- (Conv4c);
      \drawconv[5,Conv,32@$3{\times}3$,Conv5c]
      \draw (Conv4c) -- (Conv5c);
      \drawconv[6,Conv,32@$3{\times}3$,Conv6c]
      \draw (Conv5c) -- (Conv6c);
      \drawconv[7,Conv,32@$3{\times}3$,Conv7c]
      \draw (Conv6c) -- (Conv7c);
      \drawconv[8,Conv,32@$3{\times}3$,Conv8c]
      \draw (Conv7c) -- (Conv8c);
      \drawconv[9,Conv,32@$3{\times}3$,Conv9c]
      \draw (Conv8c) -- (Conv9c);
      \drawconv[10,Conv,1@$1{\times}1$,Conv10c]
      \draw (Conv9c) -- (Conv10c);
      
      \node[input node] (Out1) at (\xoffset +\xstep*11,\yoffset) {\includegraphics[width=1.25cm]{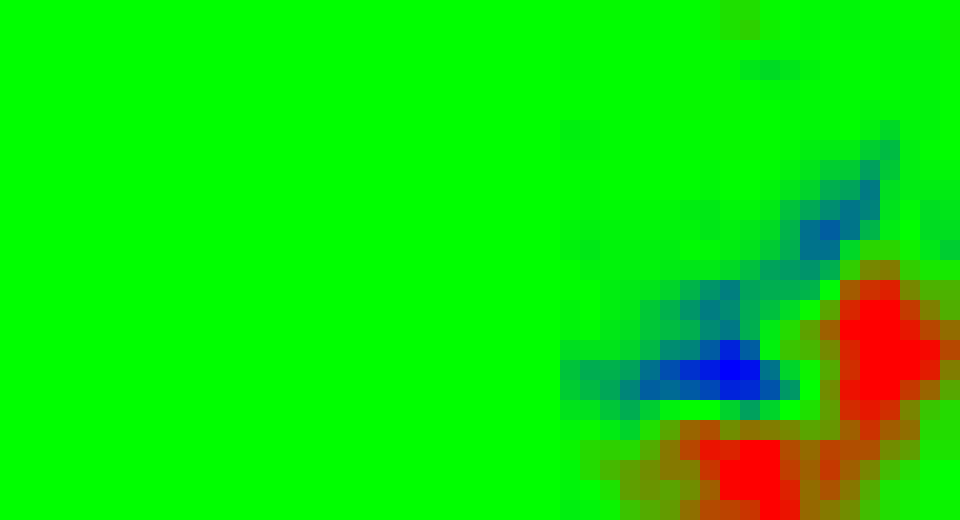}};
      \draw (Conv10c) -- (Out1);

      
      \pgfmathsetmacro{\xoffset}{1.5-0.2}
      \pgfmathsetmacro{\yoffset}{\yoffset-2.7}
      \node[elt node] (sub) at (\xoffset-0.875,\yoffset+1.2) {\parbox[top][0.4cm][c]{0.4cm}{\centering +}}; 
      \node[input node] (IniD) at (\xoffset-1.5,\yoffset+0.025) {\includegraphics[width=1.25cm]{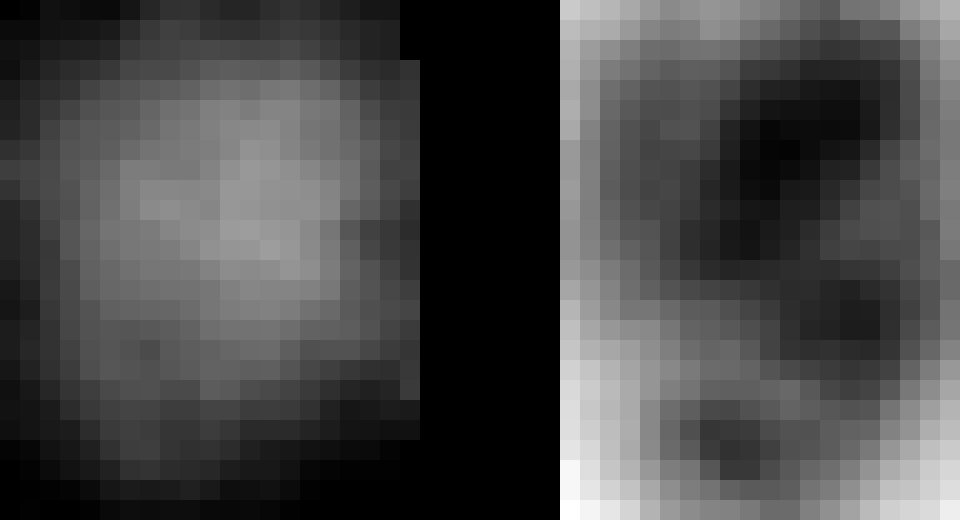}}; 
      \draw (Ing) -- (sub);
      \draw (sub) -- (IniD);
      \draw (Out1.south) to [out=270,in=0,distance=1.4cm] (sub.east);
      
      \drawconv[0,Conv,32@$3{\times}3$,Conv0iD]
      \draw (IniD) -- (Conv0iD);
      \drawconv[1,Conv,32@$3{\times}3$,Conv1iD]
      \draw (Conv0iD) -- (Conv1iD);
      \drawconv[2,Conv,32@$3{\times}3$,Conv2iD]
      \draw (Conv1iD) -- (Conv2iD);
      
      \pgfmathsetmacro{\xoffset}{\xoffset-\xstep/3}
      \pgfmathsetmacro{\yoffset}{\yoffset-0.7}
      \node[input node] (IngD) at (\xoffset-1.5,\yoffset+0.025) {\includegraphics[width=1.25cm]{action.png}}; 
      \drawconv[0,Conv,32@$3{\times}3$,Conv0gD]
      \draw (IngD) -- (Conv0gD);
      \drawconv[1,Conv,32@$3{\times}3$,Conv1gD]
      \draw (Conv0gD) -- (Conv1gD);
      \drawconv[2,Conv,32@$3{\times}3$,Conv2gD]
      \draw (Conv1gD) -- (Conv2gD);
      
      \pgfmathsetmacro{\xoffset}{\xoffset+\xstep*4}
      \pgfmathsetmacro{\yoffset}{\yoffset+0.35}
      \drawconv[0,Conv,32@$3{\times}3$,Conv0cD]
      \draw (Conv2iD) -- (Conv0cD);
      \draw (Conv2gD) -- (Conv0cD);
      \drawconv[1,Conv,32@$3{\times}3$,Conv1cD]
      \draw (Conv0cD) -- (Conv1cD);
      \drawconv[2,Conv,32@$3{\times}3$,Conv2cD]
      \draw (Conv1cD) -- (Conv2cD);
      \drawconv[3,Conv,32@$3{\times}3$,Conv3cD]
      \draw (Conv2cD) -- (Conv3cD);
      \drawconv[4,Conv,32@$3{\times}3$,Conv4cD]
      \draw (Conv3cD) -- (Conv4cD);
      \drawconv[5,Conv,32@$3{\times}3$,Conv5cD]
      \draw (Conv4cD) -- (Conv5cD);
      \drawconv[6,Conv,32@$3{\times}3$,Conv6cD]
      \draw (Conv5cD) -- (Conv6cD);
      \drawconv[7,Conv,32@$3{\times}3$,Conv7cD]
      \draw (Conv6cD) -- (Conv7cD);
      \drawconv[8,Conv,32@$3{\times}3$,Conv8cD]
      \draw (Conv7cD) -- (Conv8cD);
      \drawconv[9,Conv,32@$3{\times}3$,Conv9cD]
      \draw (Conv8cD) -- (Conv9cD);
      \drawconv[10,Conv,1@$1{\times}1$,Conv10cD]
      \draw (Conv9cD) -- (Conv10cD);
      
      \node[input node] (Out2) at (\xoffset +\xstep*11+0.2,\yoffset) {\includegraphics[width=1.25cm]{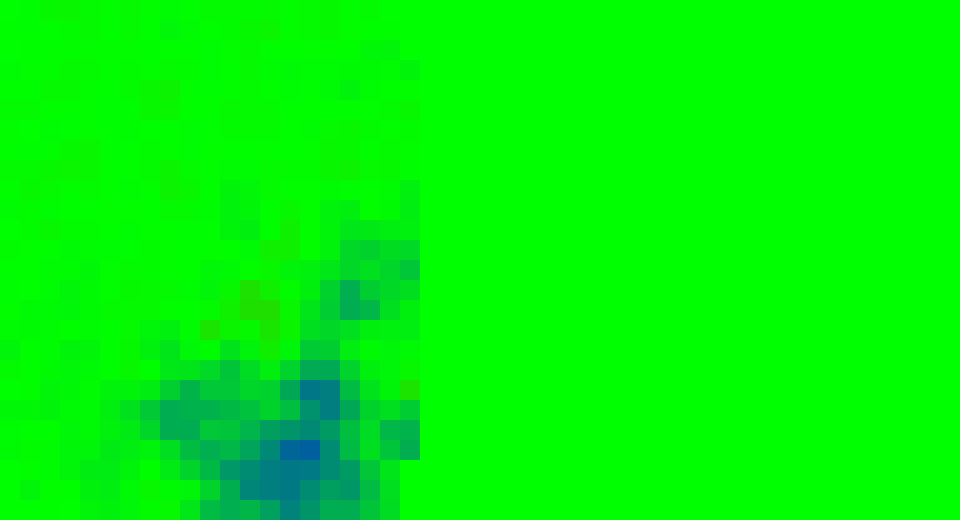}};
      \draw (Conv10cD) -- (Out2);
      
      \node[elt node] (add) at (\xoffset +\xstep*11+0.825,\yoffset+1.5) {\parbox[top][0.4cm][c]{0.4cm}{\centering +}}; 
      \draw (Out1) -- (add);
      \draw (Out2) -- (add);
      \node[input node] (Out3) at (\xoffset +\xstep*11+1.6,\yoffset+1.5) {\includegraphics[width=2cm]{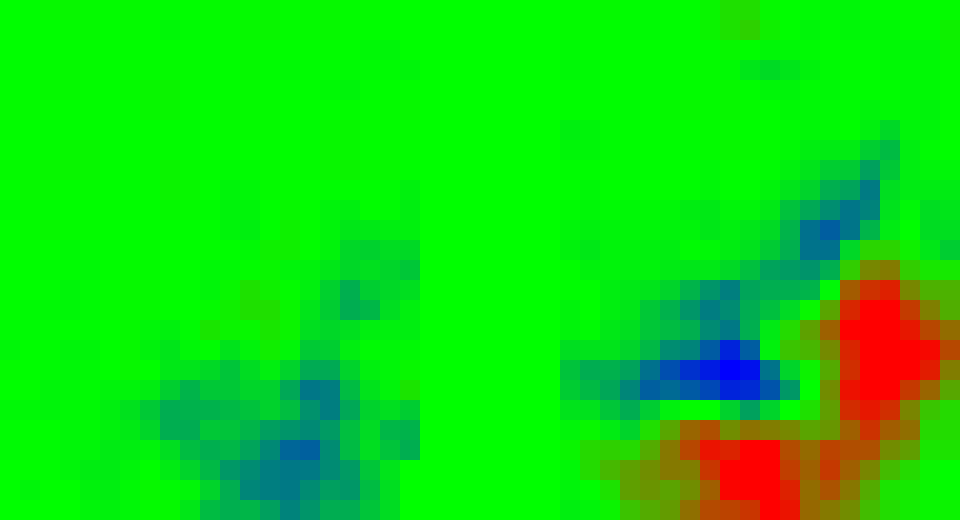}};
      \draw (add) -- (Out3);
      
    \end{tikzpicture}
\end{center}
\caption{Network layout for the {\it scoop \& dump}--net.
}
\label{fig:scoopanddump_net}
\end{figure}

The first two predictive models take the current height-map $h_t$ and a proposed action $\tilde{\theta}^i$ and attempt to predict the next state $\bar{h}_{t+1}$.
To do this the robot uses one of two fully-convolutional network architectures: the {\it single}--net and the {\it scoop \& dump}--net.
The {\it single}--net is a standard fully-convolutional network (i.e. comprised of only convolution and ReLU layers) to predict the per-grid-cell change in the height-map (the layout of the {\it single}--net is identical to the top half of Figure~\ref{fig:scoopanddump_net}).
Empirically, this network tended to have difficulty conserving mass between the {\it scoop} and {\it dump} parts of the action (i.e., the mass {\it scooped} should be approximately equal to the mass {\it dumped}).
To remedy this, we created the {\it scoop \& dump}--net, which splits the network into two halves, one to predict the change from the {\it scoop} part of the action, and the other to predict the change from the {\it dump} part of the action.
The {\it scoop \& dump}--net is shown in Figure~\ref{fig:scoopanddump_net}.
The top half of the network takes in the current height-map and the action map, and outputs the predicted change due to the {\it scoop}.
The bottom half of the network takes as input the current height-map {\it plus the changes predicted by the first network} and the action map, and predicts the changes due to the {\it dump}.
In this way, the bottom half sees what the height-map is predicted to look like after the {\it scoop} has been performed.
Additionally, to help the network enforce mass conservation, we add an explicit summation to the network that sums the change in mass due to the {\it scoop} and then passes it in a separate channel to the bottom half of the network, allowing it to directly reason about mass conservation.
Finally, we also modify the loss function for training to include an additional term for the intermediate scoop state (recorded from real-world height-maps).

\subsection{Predicting Distance to the Goal}
\label{sec:value}

\begin{figure}
\begin{center}
    \begin{tikzpicture}[->,>=stealth,auto,node distance=1.6cm,thick,
      input node/.style={rectangle,draw,anchor=west,align=center,inner sep=0,outer sep=0}]

	  \pgfmathsetmacro{\xstep}{0.5}
      \pgfmathsetmacro{\xoffset}{1.5}
      \pgfmathsetmacro{\yoffset}{3.975}
      \node[input node] (Ini) at (0.0,4.0) {\includegraphics[width=1.25cm]{before.png}}; 
      \drawconv[0,Conv,32@$3{\times}3$,Conv0i]
      \draw (Ini) -- (Conv0i);
      \drawconv[1,Conv,32@$3{\times}3$,Conv1i]
      \draw (Conv0i) -- (Conv1i);
      \drawconv[2,Conv,32@$3{\times}3$,Conv2i]
      \draw (Conv1i) -- (Conv2i);
      
      \pgfmathsetmacro{\xoffset}{\xoffset-\xstep/3}
      \pgfmathsetmacro{\yoffset}{\yoffset-0.7}
      \node[input node] (Ing) at (-0.2,3.3) {\includegraphics[width=1.25cm]{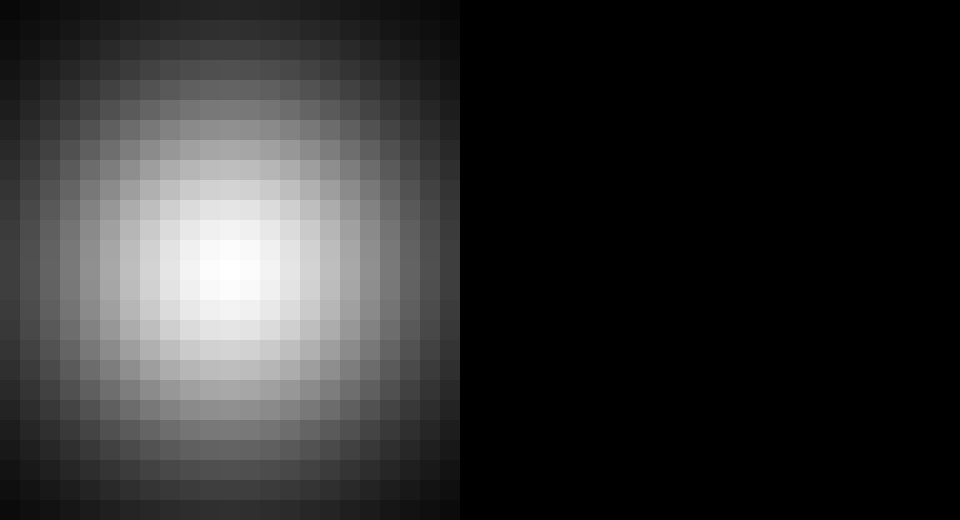}}; 
      \drawconv[0,Conv,32@$3{\times}3$,Conv0g]
      \draw (Ing) -- (Conv0g);
      \drawconv[1,Conv,32@$3{\times}3$,Conv1g]
      \draw (Conv0g) -- (Conv1g);
      \drawconv[2,Conv,32@$3{\times}3$,Conv2g]
      \draw (Conv1g) -- (Conv2g);
      
      \pgfmathsetmacro{\xoffset}{\xoffset-\xstep/3}
      \pgfmathsetmacro{\yoffset}{\yoffset-0.7}
      \node[input node] (Ina) at (-0.4,2.6) {\includegraphics[width=1.25cm]{action.png}}; 
      \drawconv[0,Conv,32@$3{\times}3$,Conv0a]
      \draw (Ina) -- (Conv0a);
      \drawconv[1,Conv,32@$3{\times}3$,Conv1a]
      \draw (Conv0a) -- (Conv1a);
      \drawconv[2,Conv,32@$3{\times}3$,Conv2a]
      \draw (Conv1a) -- (Conv2a);
      
      \pgfmathsetmacro{\xoffset}{\xoffset+\xstep*4}
      \pgfmathsetmacro{\yoffset}{\yoffset+0.7}
      \drawconv[0,Conv,32@$3{\times}3$,Conv0c]
      \draw (Conv2i) -- (Conv0c);
      \draw (Conv2g) -- (Conv0c);
      \draw (Conv2a) -- (Conv0c);
      \drawconv[1,Conv,32@$3{\times}3$,Conv1c]
      \draw (Conv0c) -- (Conv1c);
      \drawconv[2,Conv,32@$3{\times}3$,Conv2c]
      \draw (Conv1c) -- (Conv2c);
      \drawconv[3,Conv,32@$3{\times}3$,Conv3c]
      \draw (Conv2c) -- (Conv3c);
      \drawconv[4,Conv,32@$3{\times}3$,Conv4c]
      \draw (Conv3c) -- (Conv4c);
      \drawconv[5,Conv,32@$3{\times}3$,Conv5c]
      \draw (Conv4c) -- (Conv5c);
      \drawconv[6,Conv,32@$3{\times}3$,Conv6c]
      \draw (Conv5c) -- (Conv6c);
      \drawconv[7,Conv,32@$3{\times}3$,Conv7c]
      \draw (Conv6c) -- (Conv7c);
      \drawconv[8,Conv,32@$3{\times}3$,Conv8c]
      \draw (Conv7c) -- (Conv8c);
      \drawconv[9,Conv,32@$3{\times}3$,Conv9c]
      \draw (Conv8c) -- (Conv9c);
      
      \drawconv[10,Conv,16@$3{\times}3$,Conv10c]
      \draw (Conv9c) -- (Conv10c);
      \drawpool[11,AvgPool,$4{\times}4$,Pool11c]
      \draw (Conv10c) -- (Pool11c);
      \drawconv[12,Conv,16@$3{\times}3$,Conv12c]
      \draw (Pool11c) -- (Conv12c);
      \drawpool[13,AvgPool,$4{\times}4$,Pool13c]
      \draw (Conv12c) -- (Pool13c);
      
      \drawfc[14,FullyConnected,1 node,FC14c]
      \draw (Pool13c) -- (FC14c);
      
      \node[rectangle,anchor=west,align=center,inner sep=0,outer sep=0] (Out1) at (\xoffset +\xstep*16,\yoffset) {{\Huge$\mathbb{R}$}};
      \draw (FC14c) -- (Out1);
      
    \end{tikzpicture}
\end{center}
\caption{The network layout for the {\it value}--net. 
}
\label{fig:value_net}
\end{figure}

Our third predictive model attempts to infer how much closer the next state will be to the goal compared to the current state.
Specifically, given the current state $h_t$, the proposed action parameters $\tilde{\theta}^i$, and the goal state $h_g$, this model attempts to predict the following scalar quantity:
\[ \lVert h_g - \mathcal{F}\left( h_t, a_t[\tilde{\theta}^i] \right) \rVert_1 - \lVert h_g - h_t \rVert_1 \]
where $\mathcal{F}\left( h_t, a_t[\tilde{\theta}^i] \right)$ is the next state after applying action $a_t$ with parameters $\tilde{\theta}^i$.

However unlike the previous two models of Section~\ref{sec:dynamics}, this predictive model eschews explicitly computing the next state and instead uses a ConvNet to directly regress to the scalar difference in L1-norms.
The network takes as input the current height-map $h_t$, the goal height-map $h_g$, and the proposed action map $\tilde{\theta}^i$, and outputs a single real value representing its prediction for the above equation.
Intuitively this is a measure of ``value'' of a given action;
a negative value will move the state closer to the target, while a positive value will move the state away from the target.
The layout of the {\it value}--net is shown in Figure \ref{fig:value_net}.
When computing the scalar output (i.e. the ``goodness'' of a given action) the network is not required to explicitly predict the complex dynamics of the granular media.
 We compare this model to the other models in the results in section~\ref{sec:results}.

\subsection{Baseline Predictive Model}
\label{sec:baseline}

Our final predictive model is a hard-coded heuristic function as a baseline for comparison.
Similar to the first two predictive models from Section~\ref{sec:dynamics}, this model predicts an approximate next state, conditioned on the current state $h_t$ and a proposed action $\tilde{\theta}^i$.
At a high-level, the model approximates a volume of mass to be removed from the height-map at the {\it scoop} location, and places the corresponding mass directly at the {\it dump} location. The details are as follows.

Given the start pose $s^i$ and end pose $e^i$ in $\tilde{\theta}^i$, the robot draws a straight line from $s^i$ to $e^i$ over the current height map $h_t$.
Next the robot generates $h_{t + scoop}$ by removing all the granular media within $\frac{w}{2}$ cm of the line, where $w$ is the width of the scoop, with the exception of media before $s^i$ or after $e^i$.
This results in a rectangular section of $h_{t + scoop}$, running along the line segment from $s^i$ to $e^i$, set to all 0s (i.e., all the granular media removed).

The robot next computes the sum total of the volume of media removed, and divides it between $c$, the volume of media {\it scooped}, and $p$, the volume of media {\it pushed}.
For every grid cell in the rectangle affected by the {\it scoop} action described in the previous paragraph, the robot computes $\alpha_g$, the angle of the scoop.
This angle $\alpha_g$ is computed by linearly interpolating between the start angle, $\alpha_s^i$, and the end angle, $\alpha_e^i$, along the line from $s^i$ to $e^i$ (both $\alpha_s^i$ and $\alpha_e^i$ are specified by the proposed action $\tilde{\theta}^i$).
Recall from Section~\ref{sec:action} that the angle of the scoop is the angle between the plane of the tray and the plane of the scoop, with 0 degrees being parallel.
For grid cells where $\alpha_g$ is positive, their volume is added to $c$, and for grid cells where $\alpha_g$ is negative, their volume is added to $p$.
Intuitively, this means that when the open face of the scoop is facing forward (i.e., $\alpha_g$ is positive), the media is {\it scooped}, that is, it stays in the scoop when it is raised from the tray.
However, when the underside of the scoop is facing forward (i.e., $\alpha_g$ is negative), the media is {\it pushed}, that is, it is pushed in front of the scoop until it reaches $e^i$, then stays there and does not go with the scoop when it is raised.

Finally, to generate $h_{t+1}$, the robot adds two narrow Gaussians to $h_{t + scoop}$.
It adds the first at $e^i$ and adjusts it so that the volume of the Gaussian is equal to $p$.
This is the media that was {\it pushed} by the scoop.
It adds the second at the dump location specified in $\tilde{\theta}^i$ and adjusts it so that its volume is equal to $c$.
This is the media that was {\it scooped} by the scoop, and thus is deposited by the scoop at the dump location.

\section{Experimental Setup}

\subsection{Robot and Sensors}

In this paper we utilized 7 KUKA LBR IIWA robotic arms, 
however each acted independently and did not interfere with each other.
The use of multiple arms in parallel allows us to perform data collection at a faster rate.
The arms have 7 degrees of freedom and are each mounted on a horizontal table.
Another table is placed immediately in front of each robot arm and acts as the robot's workspace.
The robotic setup is shown in Figure~\ref{fig:robot}.
Each robot has it's own depth camera
capable of recording $640{\times}480$ depth images at 30 Hz.

\subsection{Objects and Granular Media}

We use pinto beans as the granular media since they are both large enough to not lodge in the robot's joints and small enough to have interesting dynamics. They also appear with minimal noise on infrared depth cameras and are rigid enough to not break or deform after extended manipulation.
For the scoop tool we use a standard large metal cooking spoon, which we rigidly attach to the robot's end-effector.
Finally, we place a tray in front of the robot and rigidly fix it to the table.
The tray we use has a divider that splits the tray into left and right halves, which allows for more interesting interactions with the granular media by the robot.
The precise pose of the tray relative to the robot and the camera are calibrated {\it a priori} and fixed for the duration of the experiments.

\subsection{Data Collection}
\label{sec:dataset}

To train the ConvNets used by the robot policy, we collected a dataset of robot scooping actions.
First, we filled one side of the tray in front of each robot with 3.75kg of pinto beans.
Next, each robot observed the tray in front of it and generated a goal state by randomly placing a target pile in one of the four corners or center of the empty side of the tray.
The robots then executed their policies, attempting to scoop the beans from the full side of the tray to the target pile on the empty side.
After 75 iterations of the {\it scoop \& dump} action, the robots randomly generated a new goal on the other side of the tray.
This process of generating goals and executing the policy was repeated until sufficient data had been collected, with the experimenter periodically resetting all the beans to one side of the tray in between repetitions.

Overall, we collected approximately 15,000 examples of the {\it scoop \& dump} action.
For the first 10,000, the robots used the baseline scoring function during policy execution as described in section~\ref{sec:baseline}.
For the last 5,000, we trained a {\it scoop \& dump}--net on the first 10,000 data-points and then used the dynamics scoring function for the policy as described in section~\ref{sec:dynamics}.

\subsection{Data Processing}

The state representation used by the policy in this paper is a height-map over the media in the tray, however the robots' observations are depth images from a depth camera.
To convert the depth images to a height-map, the robot first constructs a point-cloud from the depth image using the camera's known intrinsics.
The point-cloud is then transformed into the tray's coordinate frame and points within the tray's bounding-box are projected to the 2D discretized height-map surface, recording the resultant height above the tray.
Figures \ref{fig:depth} and \ref{fig:height_map} show a depth image and its corresponding height-map.
The robot's arm and the scoop are excluded from the image by moving the arm up and out of the robot's view.

\section{Evaluation}


\subsection{Robot Policy}
\label{sec:policy}

To select the correct actions, the robot utilizes a greedy policy based on the cross-entropy method (CEM) \cite{de2005}. 
CEM is a sampling method that uses a scoring function $L$ to iteratively sort samples, estimate a distribution from the top $K$, and resample.
To compute the action parameters $\theta_t$, the robot first samples $N$ sets of parameters $\tilde{\theta}^1, ..., \tilde{\theta}^N$ uniformly at random from the parameter space.
Next, for each sampled parameter set $\tilde{\theta}^i$, the robot computes the score $L(\tilde{\theta}^i, h_t, h_g)$ using the scoring function $L$, the current state $h_t$, and the goal state $h_g$.
Intuitively, $L$ is a measure of how ``good'' the robot believes taking an action with parameters $\tilde{\theta}^i$ in state $h_t$ towards goal $h_g$ will be, with lower values indicating better parameters.
The robot then discards all but the best $K$ sets (i.e., the lowest scoring $K$ sets), where $K < N$.
Next the robot fits a Gaussian distribution to the $K$ samples, and then resamples $N$ new sets of parameters from this Gaussian.
The robot repeats this loop of sampling from a distribution, picking the best $K$, and refitting the distribution, for a fixed number of steps.
On the final iteration, the robot sets $\theta_t$ to be the values of the top performing set of parameters $\tilde{\theta}^i$.

The scoring function $L$ uses the predictive model underneath to compute the score.
For the models in sections \ref{sec:dynamics} and \ref{sec:baseline}, since they both predict the next state, the scoring function simply returns the L1 error (mean absolute error) between the predicted next state and the goal state.
For the model described in section~\ref{sec:value}, the robot directly uses the output of the network as the score since it already predicts how ``good'' a proposed action is.
For the results in section~\ref{sec:results}, we compare using each of the predictive models in the scoring function.

\subsection{Training the Models}
The predictive models described in sections \ref{sec:dynamics} and \ref{sec:value} utilize ConvNets.
To train those networks, we use the dataset collected in section~\ref{sec:dataset}.
For all networks, we use the mini-batched gradient descent method Adam \cite{kingma2014} with a learning rate of $5\mathrm{e}{-4}$ to iteratively update the weights of the networks.
All weights were initialized randomly.
We first pre-trained all the networks for 30,000 iterations using the baseline model output predictions (using initial states from the robot data) described in section~\ref{sec:baseline}\footnote{We empirically determined that the networks performed better with pre-training than without.}.
Next we trained the networks for 100,000 iterations on the initial and next states captured from the robot.

Since each network was slightly different from the others, we used different loss functions for each.
For the {\it single}--net (described in section~\ref{sec:dynamics}), we use the L2-loss on the predicted next state.
For the {\it scoop \& dump}--net (also described in section~\ref{sec:dynamics}), we also use the L2-loss on the predicted next state, but we add another L2-loss on the output of the top half of the network (the {\it scoop} only prediction), which is added with equal weight.
Additionally we stop the gradients from the first loss from propagating directly back into the top half of the network, which encourages the top half to predict only the result of the {\it scoop} part of the action.
Finally, for the {\it value}--net (described in section~\ref{sec:value}) we use an L2-loss on the change in mean absolute error to that goal from the before action state to the after action state (each scoop action in the dataset had a corresponding goal).

\subsection{Tasks}

We evaluate the robot's models on two example tasks, a ``simple'' task and a ``hard'' task.
The first task is for the robot to scoop on one side of the tray that is filled with beans, and dump the beans into a Gaussian-shaped pile on the empty side.
The robot is given 100 {\it scoop \& dump} actions to complete the goal shape.
While this may seem relatively simple, it is actually quite complicated.
The robot must be able to reason about what types of scoop actions will result in beans collecting in the scoop. It also must be able to reason about how many beans it will acquire and about where those beans will go when dumping them.
That is, for even such a ``simple'' task the robot must have an intimate understanding of the dynamics of the granular media.
We also evaluate the robot on a ``hard'' task that adds another level of complexity.
In this task, the robot must dump the beans into a more complex shape, requiring the robot to reason in more detail about the dynamics of the dumped beans.
In this paper, we use the shape of the letter ``G'' (in Baskerville typeface) as the target.

\section{Results}
\label{sec:results}

\setlength{\objectsize}{5.0cm}
\begin{figure}
    \setlength{\fboxsep}{0pt}
    \setlength{\fboxrule}{1pt}
    \setlength{\unitlength}{1.0cm}
    \hspace{-0.5cm}%
    \begin{subfigure}[b]{\objectsize}
        \hspace{-0.2cm}\includegraphics[width=\objectsize]{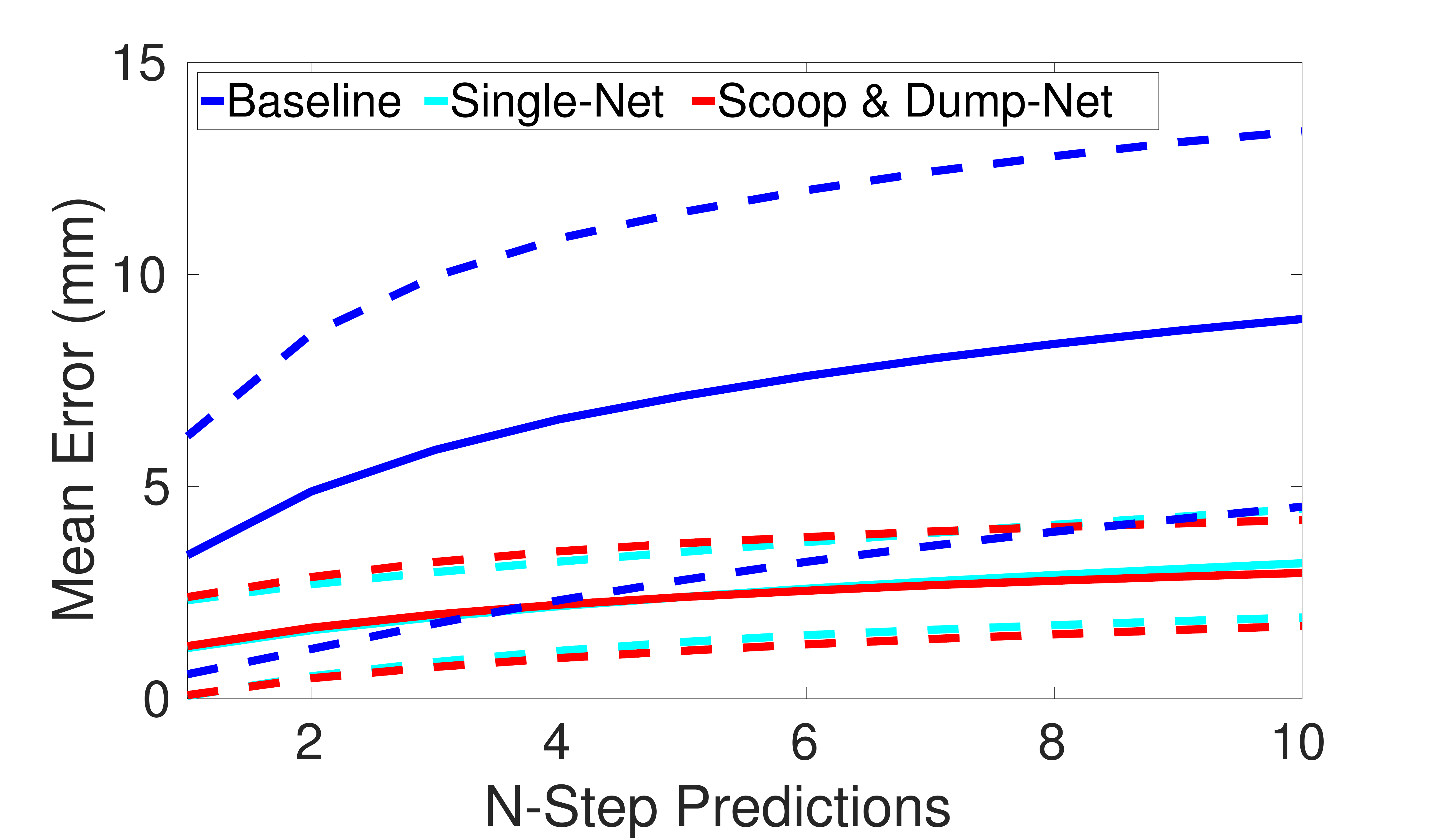}
        \caption{Test Set Error}
        \label{fig:mse}
    \end{subfigure}%
    \begin{subfigure}[b]{\objectsize}
        \hspace{-0.2cm}\includegraphics[width=\objectsize]{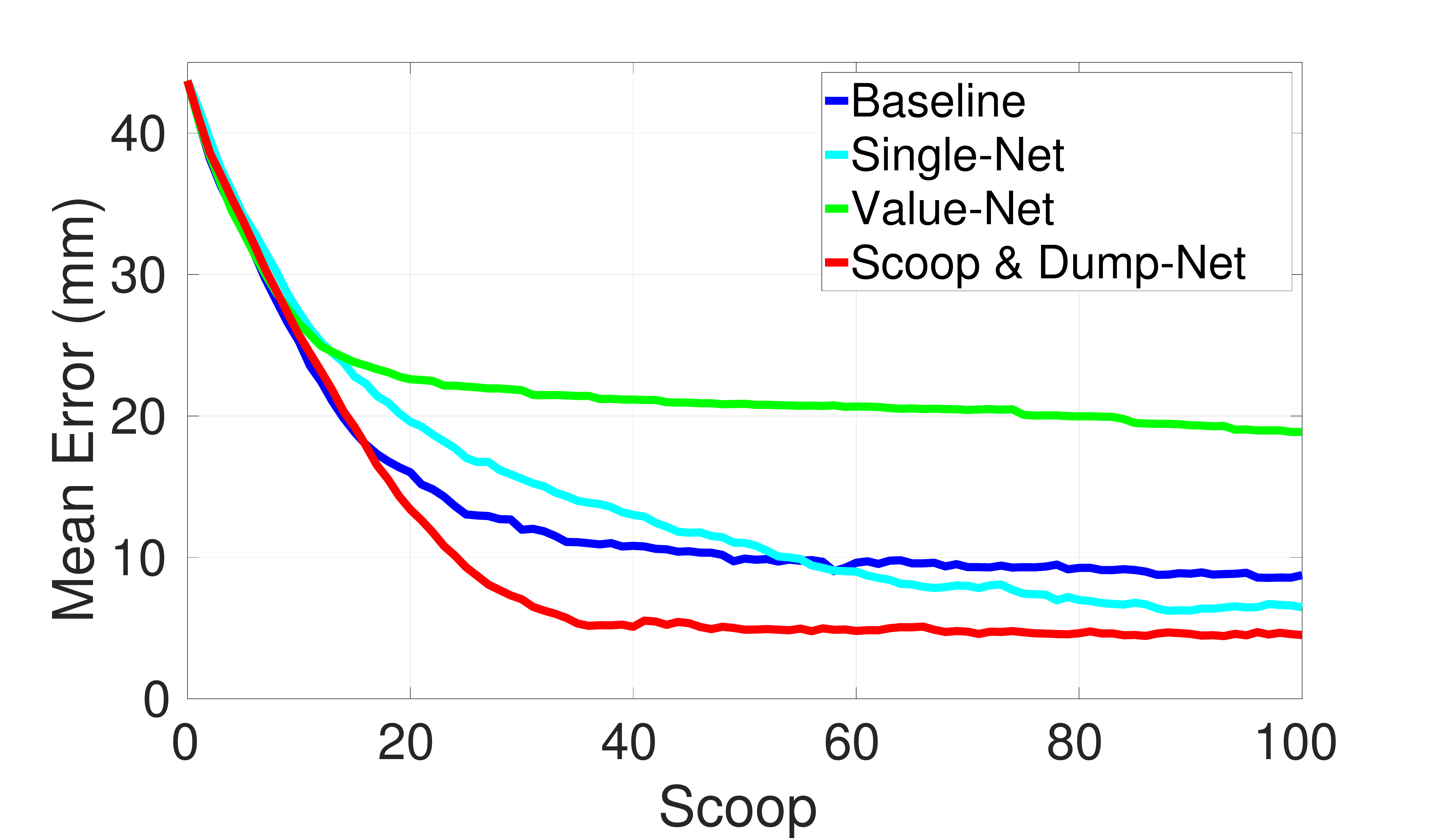}
        \caption{Pile Task Error}
        \label{fig:pile_plot}
    \end{subfigure}%
    \begin{subfigure}[b]{\objectsize}
        \hspace{-0.2cm}\includegraphics[width=\objectsize]{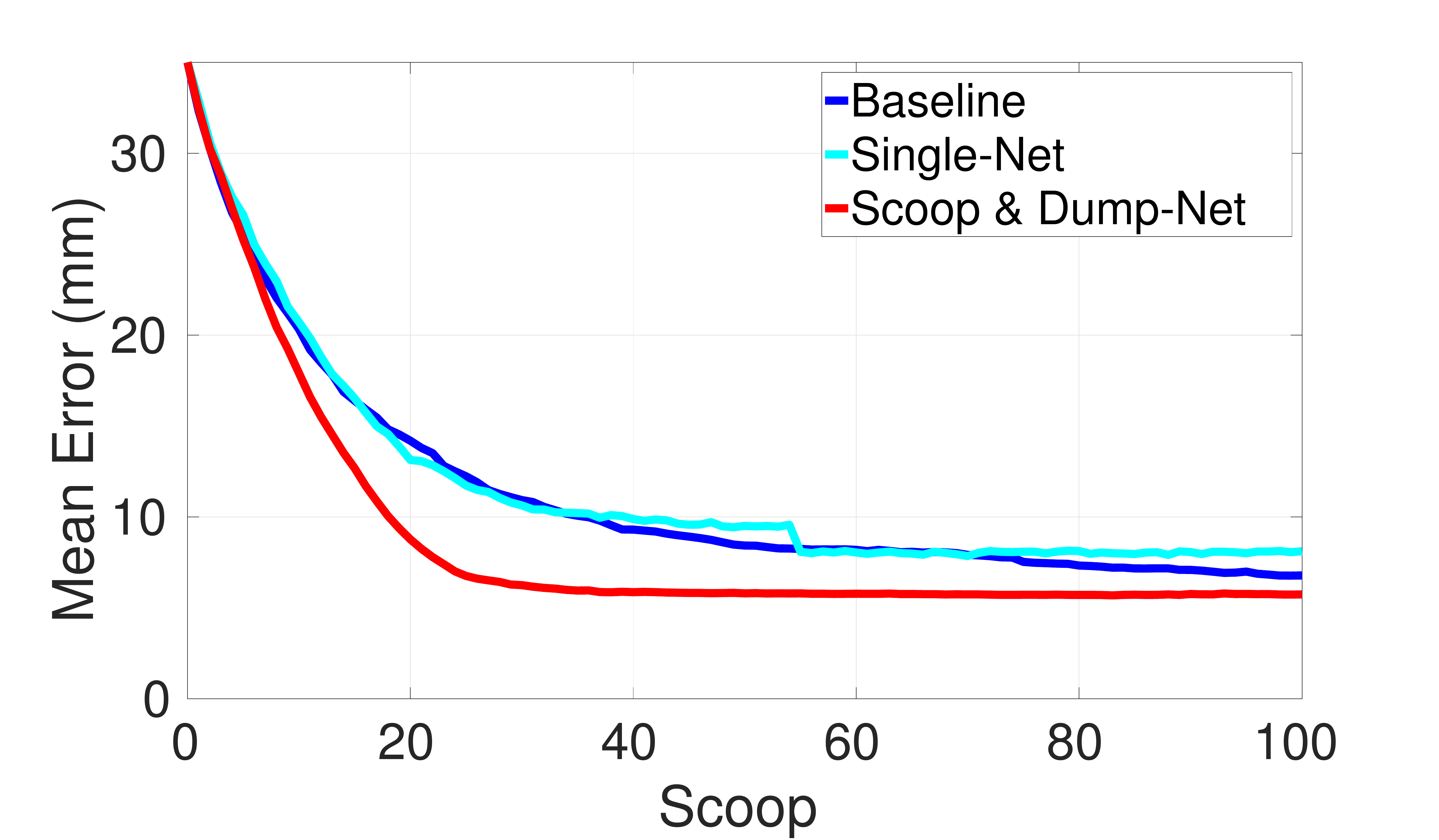}
        \caption{G Task Error}
        \label{fig:g_plot}
    \end{subfigure}
    \caption{The results of evaluating our predictive models. Left to right: mean error (in mm, confidence regions indicated by dashed lines) of the 3 models predicting the state transition; the mean error of the 4 models on the pile task and on the G task.}
    \label{fig:results}
\end{figure}


Figure~\ref{fig:results} shows the results\footnote{Please refer to the associated video for additional results: \url{https://youtu.be/YUb5hi-OK70}}.
Figure~\ref{fig:mse} shows the error between predicted next states and the actual next states on a test set for each of the three models that predict state transitions \footnote{For this portion of the analysis we don't compare with the {\it value}--net because there is no next-state prediction to compare with, although empirically the error for it did seem to converge to a reasonable value after training.}.
The X-axis shows how the models perform over multi-step predictions (i.e., using their own output as input for the next timestep), and the Y-axis shows the average error in millimeters, with the dashed lines indicating the standard deviation for each model.
Interestingly, the {\it single}--net and the {\it scoop \& dump}--net perform very similarly on our test set.
As expected, the rough heuristic baseline model performs poorly as it does not attempt to accurately predict the dynamics but only roughly captures the overall behavior.

From Figure~\ref{fig:pile_plot} it is apparent that while the baseline does not predict accurate dynamics, it is never-the-less able to solve the pile task to some degree.
Figure~\ref{fig:pile_plot} shows the error in millimeters for each model averaged across 3 runs for the pile task with each run initialized to the same mass of beans (3.75kg) and leveled in the same container.
The X-axis shows the error over time as the robot completes more {\it scoop \& dump} actions (100 total).
In all cases the error goes down over time as expected.
From the graph it is apparent that the {\it value}--net performs much more poorly than the other models and is unable to reason effectively about the robot's actions.
Without explicitly training the network to learn the rich, high-dimensional statistics of the full output state representation (as in the {\it scoop \& dump}--net), it may easily get stuck in local minima during training (particularly in the regime of limited training data) and as a result performs poorly.
It is clear that it is unsuitable for these kinds of tasks and so we leave it out of the remaining evaluations.
Interestingly, the baseline method outperforms the {\it single}--net method for the first 50 of 100 actions, and although it does eventually converge to a state closer to the goal, it does so much more slowly.
The {\it scoop \& dump}--net, on the other hand, not only reaches a state closer to the goal, but converges much faster than any other method.
This clearly indicates that even though the {\it single}--net and {\it scoop \& dump}--net had similar performance on the test set, when actually using the models on control tasks, the {\it scoop \& dump}--net is better able to inform the policy about the dynamics of the actions on the granular media.

Examining Figure~\ref{fig:g_plot}, it is clear that this trend remains true for even more complicated tasks.
This figure shows the performance of 3 of the models on the G task
(we left out the {\it value}--net due to its poor performance on the pile task) 
averaged across three runs each.
Interestingly, the baseline, while converging more slowly in this case than the {\it single}--net, in the end outperforms it by a narrow margin.
These results taken together indicate that the structure of the {\it scoop \& dump}--net is better suited to acting as a predictive model of a robotic control policy involving granular media than a naive unstructured network like the {\it single}--net.

\vspace{-0.1cm}
\section{Conclusion}
\vspace{-0.1cm}

In this paper, we developed 4 predictive models for enabling a robotic control policy to manipulate granular media into a desired shape.
We evaluated these models on both a test set and on a pair of manipulation tasks of varying difficulty.
The results clearly show that for tasks involving granular media, the best performance can be achieved by training structured ConvNets to predict full state transitions. 
Our {\it scoop \& dump}--net, which explicitly modeled the different parts of the action in the network structure, outperformed both our {\it value}--net, which only predicted the value of an action with respect to a goal, and our {\it single}--net, which predicted the full state transition but did not have any internal structuring to reflect the nature of the action.
This is consistent with prior work \cite{byravan2016se3}, which showed that structured networks are also necessary when predicting the motion of dense point clouds.
Our results show using structured, learned predictive models can enable robots to reason about substances such as granular media that are difficult to reason about with traditional kinematic models.




\acknowledgments{The authors would like to thank Peter Pastor for his invaluable insights and patient help.}


\bibliography{references}  

\end{document}